\documentclass[11pt]{article}

\usepackage[margin=1in]{geometry}
\usepackage{graphicx}
\usepackage{float}
\usepackage{booktabs}
\usepackage{amsmath}
\usepackage{hyperref}

\title{Does the Competitive Component of Adversarial Self-Play Improve Legal Reasoning? A Controlled Negative Result}
\author{Miseong Shawn Kim}
\date{Preprint, 2026-08-02}

\usepackage{adjustbox}
\usepackage{pgfplots}
\pgfplotsset{compat=1.17}
\begin{document}

\maketitle

\begin{abstract}
Adversarial self-play is an appealing recipe for legal reasoning: have a student
model draft an argument, have an adversary attack it, and reward the student when
its argument \emph{survives} the attack. We designed exactly such a training signal ---
a verifiable ``survival'' reward in which both the student's cited authorities and
the adversary's counter-authorities are checked by a citation verifier, so that
survival is decided on verified grounds rather than rhetoric, and fabricated
citations are automatically neutralized. We then asked a narrow but important
question: \textbf{does the competitive component itself --- the adversary and the
survival reward --- add anything on top of an otherwise identical non-competitive
training run?} Across four independent tests --- a bootstrap comparison, a
two-seed replication, a paired per-case adversarial-robustness comparison, and a
blinded head-to-head judgment of generated arguments, plus a follow-up pilot with
a deliberately strengthened self-play adversary --- the competitive component
produced \textbf{no reliable benefit}. The blinded judgment gave a 49\% win rate
(binomial $p \approx 1.000$); the strengthened-adversary pilot gave a 50\% win rate
(32:32, $p \approx 1.000$). An early apparent $+29\%$ advantage reversed and proved to be a
small-sample artifact. We report this as an honest negative result. The value of
the paper is reproducibility and the sharing of concrete pitfalls: an
initially-promising metric that inverted on more data, and an
adversarial-robustness metric that silently collapsed to plain recall once the
adversary stopped citing the same authorities as the gold answer. This null is consistent with, and reconfirms in the legal domain, the conclusion of the companion coding-domain study (Kim, 2026, arXiv:2607.08255) that the value of multi-teacher curricula arises from constructing a verifiable environment rather than from competition itself.
\end{abstract}

\section{Introduction}

The intuition behind adversarial self-play for legal reasoning is strong. A brief
that can withstand a competent opponent's rebuttal is presumably better than one
that cannot, so training a student to \emph{survive} an adversary's attack should teach
robust reasoning. This intuition motivated us to build a verifiable survival
reward and to fold it into a reinforcement-learning-from-verifiable-rewards
(RLVR) curriculum for legal argument generation.

But the intuition conflates two things: (i) exposure to hard, adversarially
selected cases, and (ii) the \emph{competitive} mechanism itself --- the live adversary
and the survival reward. A training run can benefit from harder data without the
competition adding anything. The clean scientific question is whether the
competitive component, holding the rest of the curriculum fixed, improves legal
reasoning.

We ran controlled comparisons to answer exactly that question and obtained a null
result: the competitive component did not reliably help. We publish this because
negative results in this space are under-reported, because our measurement
pitfalls are instructive, and because the null was robust across four
methodologically different tests plus a follow-up pilot that specifically
targeted the most plausible confound (a weak adversary).

Our contribution is therefore not a method that works but a \textbf{credible negative
result with a documented experimental design and two concrete measurement
traps}, so that others do not spend the same effort rediscovering them.

\section{Related Work}

\textbf{ASP2LJ --- An Adversarial Self-Play Lawyer Augmented Legal Judgment Framework} (arXiv:2506.18768,
\url{https://arxiv.org/abs/2506.18768}) proposes adversarial self-play for legal
judgment. Our study is complementary and cautionary: we isolate the competitive
component and find it does not, in our setting, add value beyond the shared
curriculum.

\textbf{AgentCourt --- Simulating Court with Adversarial Evolvable Lawyer Agents} (ACL
Findings 2025, \url{https://aclanthology.org/2025.findings-acl.304/}) shows evolvable
adversarial lawyer agents improving over cases. We differ by measuring, with
controls, whether the competitive signal transfers into a trained model's
reasoning quality.

\textbf{LegalSim --- Multi-Agent Simulation of Legal Proceedings} (arXiv:2510.03405,
\url{https://arxiv.org/abs/2510.03405}) simulates proceedings with multiple agents. Our
work asks a downstream training question rather than a simulation-fidelity
question.

\textbf{Courtroom-style multi-agent debate} (arXiv:2603.28488,
\url{https://arxiv.org/abs/2603.28488}) verifies controversial claims via progressive RAG and role-switching. We test
whether turning such adversarial interaction into a training reward yields a net
gain over a non-competitive baseline, and report that it did not in our
experiments.

\textbf{Broader self-play, debate, and adversarial training.} Beyond the legal
setting, the competitive-training intuition has deep roots. \emph{AI Safety via Debate}
(Irving et al., 2018) frames competition as a mechanism to surface weak arguments;
\emph{Self-Play Fine-Tuning} (SPIN; Chen et al., ICML 2024) shows self-generated play
can strengthen weak models \emph{without a live competitor}; and \emph{Constitutional AI}
(Bai et al., 2022) obtains gains from non-competitive self-critique. \emph{Adversarial
NLI} (Nie et al., ACL 2020) demonstrates the value of human-model adversarial data
collection. Crucially, a parallel line --- \emph{Chain-of-Thought Prompting} (Wei et al.,
2022) and \emph{Self-Consistency} (Wang et al., ICLR 2023) --- shows that much of the gain
attributed to ``reasoning'' methods actually comes from added reasoning \emph{structure}
or \emph{sampling}, not from competition per se. Our contribution is exactly this
isolation: holding data, model, and compute fixed, we test whether the competitive
component alone adds anything, and find it does not in our setting. To our
knowledge this competitive-vs-noncompetitive ablation has not been reported for
legal reasoning.

On the general point that reward design in verifiable-reward RL can silently
degenerate --- a reward that appears to measure the target but is satisfied by a
proxy --- our second measurement trap (Section 5) is a concrete legal-domain
instance.

The author's companion study in the coding domain (Kim, 2026, arXiv:2607.08255)
shows that the value of a multi-teacher ``compete then collaborate'' curriculum
comes not from answer pooling or from competition itself but from \textbf{building a
verifiable environment}, which is consistent with this paper's negative result on
the competitive component.

\section{Method}

\subsection{The adversarial curriculum and the survival reward}

We built an adversarial case-pack curriculum in which each training case carries
an adversarial layer: an attack from an opposing role (arguing element failure,
precedent distinction, evidence exclusion, procedural defect), a weakness
diagnosis, a defense strategy, the student's argumentation, a judge simulation,
and rebuttal rounds. Roles were separated to avoid self-bias (attacker and judge
were distinct from the student), and all attack and defense citations were
required to pass a citation verifier.

The reinforcement signal combined immediate verifiable rewards --- citation
correctness, legal-element coverage, and output-schema conformance --- with a
\textbf{survival reward}: the student drafts an argument, an adversary rebuts it, and
survival is scored on \emph{verified} grounds. Concretely, both the student's cited
authorities and the adversary's counter-authorities are passed through the
citation verifier, so a duel is decided on verifiable authority rather than
fluent prose, and a fabricated citation scores zero on the citation reward and
therefore cannot win a duel by bluffing. Self-play used a frozen copy of the
student as the adversary, refreshed periodically to the latest checkpoint.

We follow the finding from the broader RLVR literature that imitation-style SFT
can \emph{degrade} an already-capable student while the same curriculum delivered as
RLVR can improve it; this is why the survival signal was posed as a verifiable
reward rather than as supervised imitation.

\subsection{The controlled question and the comparison design}

The competitive component is the \emph{only} thing we vary. The treatment run uses the
live adversary and the survival reward; the control run uses an otherwise
identical setup without the competitive component (a non-competitive,
verifiable-reward run of comparable compute). We then compared them four ways:

\begin{enumerate}
\item \textbf{Bootstrap comparison} of the primary held-out reasoning metric.
\item \textbf{Two-seed replication} (two independent seed pairs) of the same comparison.
\item \textbf{Paired per-case adversarial robustness}: for a fixed set of cases, measure
   each model's ``retained'' score after an adversary attack, paired case by case,
   and test the difference.
\item \textbf{Blinded head-to-head judgment}: an independent judge model, given generated
   arguments from the two systems in randomized order (position-bias guarded),
   picks a winner; test with a binomial.
\end{enumerate}

Finally, because the most plausible explanation for a null is a \emph{weak} adversary,
we ran a follow-up pilot that deliberately \textbf{strengthened} the self-play
adversary: a genuine multi-round self-play loop in which the adversary was
refreshed to the latest student checkpoint each round, then a blinded
head-to-head against a non-competitive baseline of matched compute.

\section{Experiments}

All numbers below are our own measurements on our data and models. Sample sizes
are small and are stated explicitly; we do not round them up into stronger claims
than they support.

\textbf{Test 1 --- Bootstrap.} The competitive run's advantage on the primary held-out
metric was not significant (bootstrap; the effect was small and its confidence
interval included zero).

\textbf{Test 2 --- Two-seed replication.} Across two seed pairs, the competitive run
showed a \emph{directionally} positive but individually non-significant F1 difference
(seed-42 $\Delta \approx +0.0042$, seed-7 $\Delta \approx +0.0068$). We initially recorded this as ``a
small but reproducible positive direction,'' which --- as Test 3 shows --- was not
robust.

\textbf{Test 3 --- Paired per-case adversarial robustness (the reversal).} On an initial
$n=18$ adversarial set, the competitive run appeared to have a large advantage in
retained robustness (about $+29\%$). When we expanded the paired set to $n=29$, the
sign \textbf{reversed}: $\Delta$(competitive $-$ control) retained $\approx -0.0069$, CI $\approx [-0.021, 0]$,
non-significant, with 28 of 29 cases tied, and competitive $0.051 <$ control $0.058$.
The initial $+29\%$ was a small-sample artifact. This is the paper's first pitfall:
a headline effect from $n=18$ that did not survive $n=29$.

\textbf{Test 4 --- Blinded head-to-head judgment.} An independent judge, comparing
generated arguments in randomized order, returned competitive 28 : control 29 :
tie 9 --- a \textbf{49\% win rate, binomial $p \approx 1.000$}. Every test to this point
converged on null.

\textbf{Follow-up pilot --- strengthened adversary.} To rule out ``the adversary was too
weak,'' we ran a genuine multi-round self-play pilot in which the adversary was
refreshed to the latest student checkpoint each round (verified-citation strength
of the adversary rising across rounds, e.g. $56 \rightarrow 62$ verified citations), then a
blinded head-to-head against an F1-only (non-competitive) baseline of matched
compute. Result: \textbf{32 : 32, 2 ties, 50\% win rate, $p \approx 1.000$.} The ``weak
adversary'' hypothesis is rejected: even with a strengthened self-play adversary,
the competitive component added no reliable benefit.

\textbf{Aggregate.} Bootstrap ($\sim$0.41), two-seed (small, individually non-significant,
and not robust), paired A-Bench (reversed to null on more data), blinded judgment
(28:29, $p \approx 1.000$), and the strengthened-adversary pilot (32:32, $p \approx 1.000$) all
converge: \textbf{the competitive component of adversarial self-play did not, in our
controlled experiments, improve legal reasoning.}

We emphasize the small samples ($n=18 \rightarrow 29$ for the paired test; $\sim$66 held-out
arguments for the judgment; tens of duels in the pilot). These are pilot-scale
experiments, not large-N confirmations. The consistency of the null across four
different methodologies is what we lean on, not the power of any single test.

\section{Pitfalls and Lessons (the reproducibility contribution)}

\textbf{Pitfall 1 --- the $n=18$ mirage.} A $+29\%$ retained-robustness advantage at $n=18$
reversed to a slight \emph{deficit} at $n=29$. Small adversarial sets can manufacture
large, plausible-looking effects. Expanding the sample before believing a
headline is not optional here; it is the difference between a positive and a
negative conclusion.

\textbf{Pitfall 2 --- the reward that collapsed to recall.} Our survival/retained metric
was built on citation overlap between the student's argument and gold
authorities. When we inspected the adversary, we found that a \emph{strong} opposing
advocate does not cite the gold authorities --- it cites the \emph{opposing} law to make
its case. As a result the rebuttal term of the survival reward was effectively
inert, and ``retained'' degenerated into plain F1 recall. An earlier adversary that
appeared to hit gold authorities $\sim$0.32 of the time was doing so only because it
was drifting out of its adversarial role. This is a concrete legal-domain example
of a verifiable-reward metric silently measuring the wrong thing; we redesigned
the final judgment to a blinded win/lose comparison precisely to escape this
trap.

\textbf{General lesson.} Early, cheap probes (we caught a \verb|gold_hit = 0| on a
two-case manual test) prevented days of wasted training on a defective metric.
Verifiable-reward pipelines need adversarial audits of the \emph{reward itself}, not
just of the model.

\subsection*{5.1 Limitations and Statistical Power (honest scope)}

This is a \textbf{pilot-scale, underpowered} study, and we state that plainly rather
than dress it as a definitive refutation. The paired robustness test moved from
$n=18$ to $n=29$; the blinded judgment used $\sim$66 arguments; the strengthened-adversary
pilot compared tens of duels. With these sizes we can credibly say \emph{``we found no
reliable benefit across four different tests,''} but we \textbf{cannot} bound the true
effect tightly enough to assert it is exactly zero. The most likely residual
confounds are (i) our specific adversary and reward instantiation, (ii) a single
task domain, and (iii) limited seeds. The standard reviewer objection to a
negative result --- \emph{``your setup was simply too weak''} --- is partly pre-empted by the
strengthened-adversary pilot, but not eliminated.

\textbf{Toward a definitive result.} Before this becomes more than a cautionary pilot,
the controlled comparison should be re-run at higher power: identical
competitive/non-competitive conditions, $\geq 5$ seeds with a pre-registered primary
metric, an explicit power analysis, bootstrap CIs on the paired robustness metric,
and at least one second task domain. We therefore present the current result as a preliminary, small-sample finding;
a higher-powered replication is needed to establish a definitive effect size.
The paper's durable value --- independent of effect size
--- is the two documented measurement traps (the $n=18$ mirage and the survival reward
that collapsed to recall), which are reproducibility contributions in their own
right.

\section{Conclusion}

We designed a verifiable survival reward for adversarial self-play in legal
reasoning and asked, under controls, whether the competitive component adds value
over a matched non-competitive baseline. Four methodologically distinct tests
plus a strengthened-adversary pilot returned a robust null (win rates 49\% and
50\%, $p \approx 1.000$). An early $+29\%$ advantage was a small-sample artifact, and our
survival metric had silently collapsed to recall because strong adversaries cite
opposing rather than gold authority. We publish this negative result, with its
design and its two pitfalls, so that the field can price the competitive
component honestly and avoid the same measurement traps. We do not claim the
competitive component \emph{can never} help --- only that, in our controlled,
small-sample setting, it did not, and that the mechanism designed to make it help
was itself the first thing that needed auditing. This null is consistent with, and reconfirms in the legal domain, the conclusion of the companion coding-domain study (Kim, 2026, arXiv:2607.08255) that the value of multi-teacher curricula arises from constructing a verifiable environment rather than from competition itself.

\section*{Results Summary (our measurements)}

\begin{table}[h]
\centering
\begin{adjustbox}{max width=\textwidth}
\begin{tabular}{llll}
\toprule
Test & Metric & Result & Significance \\
\midrule
Bootstrap & primary held-out metric & small effect, CI includes 0 & n.s. ($\sim$0.41) \\
Two-seed & F1 $\Delta$ & seed-42 $+0.0042$ / seed-7 $+0.0068$ & individually n.s. \\
Paired robustness & retained $\Delta$ (comp$-$ctrl) & $n=18$ $+29\%$ $\rightarrow$ \textbf{reversed at $n=29$: $-0.0069$} & n.s. (28/29 tied) \\
Blinded judgment & win:lose:tie & 28 : 29 : 9 (\textbf{49\%}) & binomial $p\approx1.000$ \\
Strengthened-adversary pilot & win rate & 32 : 32 (\textbf{50\%}) & $p\approx1.000$ \\
\bottomrule
\end{tabular}
\end{adjustbox}
\end{table}

\begin{figure}[H]\centering
\begin{tikzpicture}
\begin{axis}[xbar,bar width=12pt,width=0.8\linewidth,height=4cm,xmin=0,xmax=100,
symbolic y coords={blinded,pilot},ytick=data,xlabel={competitive win rate (\%)},
nodes near coords,every node near coord/.append style={font=\footnotesize}]
\addplot coordinates {(49,blinded) (50,pilot)};
\end{axis}\end{tikzpicture}
\caption{Competitive-component win rate: 49\% (blinded, 28:29:9) and 50\% (strengthened-adversary pilot, 32:32), both $p\approx1.000$. The 50\% mark = no benefit; the $+29\%$ at $n{=}18$ reversed at $n{=}29$. All four tests converge to null.}
\label{fig:r4}
\end{figure}
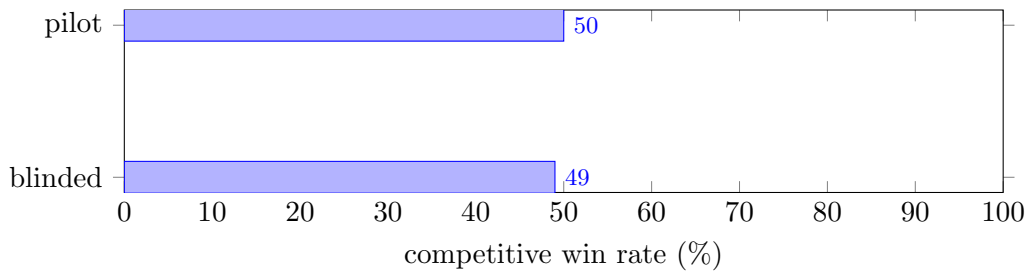

\section*{References}

\begin{enumerate}
\item ASP2LJ: An Adversarial Self-Play Lawyer Augmented Legal Judgment Framework. arXiv:2506.18768.
   \url{https://arxiv.org/abs/2506.18768}
\item AgentCourt: Simulating Court with Adversarial Evolvable Lawyer Agents. ACL
   Findings 2025. \url{https://aclanthology.org/2025.findings-acl.304/}
\item LegalSim: Multi-Agent Simulation of Legal Proceedings. arXiv:2510.03405.
   \url{https://arxiv.org/abs/2510.03405}
\item Courtroom-Style Multi-Agent Debate with Progressive RAG and Role-Switching for Controversial Claim Verification. arXiv:2603.28488.
   \url{https://arxiv.org/abs/2603.28488}
\item Irving, Christiano, Amodei. AI Safety via Debate. arXiv:1805.00899.
\item Chen et al. Self-Play Fine-Tuning Converts Weak Language Models to Strong Reasoners (SPIN). ICML 2024.
\item Bai et al. Constitutional AI: Harmlessness from AI Feedback. arXiv:2212.08073.
\item Nie et al. Adversarial NLI: A New Benchmark for Natural Language Understanding. ACL 2020.
\item Wei et al. Chain-of-Thought Prompting Elicits Reasoning in Large Language Models. NeurIPS 2022.
\item Wang et al. Self-Consistency Improves Chain of Thought Reasoning in Language Models. ICLR 2023.
\item Kim, Miseong Shawn. Compete Then Collaborate: Frontier AI Teachers Build a Verifiable Curriculum to Improve a Coding Student Beyond Imitation. arXiv:2607.08255 (cs.AI), 2026. \url{https://arxiv.org/abs/2607.08255}
\end{enumerate}

\end{document}